\documentclass{article}
\usepackage{amsmath,epsfig}
\usepackage[preprint]{spconfa4}
\usepackage{xcolor}
\usepackage{cite}
\usepackage{booktabs}
\usepackage{arydshln}
\usepackage{multirow,subfigure}
\usepackage{amssymb}
\usepackage{bbding}
\usepackage{color}
\usepackage{bbding}
\usepackage{pifont}
\usepackage{fancyhdr}

\let\OLDthebibliography\thebibliography
\renewcommand\thebibliography[1]{
  \OLDthebibliography{#1}
  \setlength{\parskip}{0pt}
  \setlength{\itemsep}{0pt plus 0.3ex}
}

\begin{document}\sloppy

\def\x{{\mathbf x}}
\def\L{{\cal L}}

\title{THQA: A Perceptual Quality Assessment Database for Talking Heads}
%

\name{%
\begin{tabular}{@{}c@{}}
Yingjie Zhou$^{1,2}$ \qquad 
Zicheng Zhang$^{1,2}$ \qquad 
Wei Sun$^{1,2}$ \qquad 
Xiaohong Liu$^{1,2}$ \qquad  
Xiongkuo Min$^{1,2}$ \\
Zhihua Wang$^{1,2}$ \qquad
Xiao-Ping Zhang$^{3}$, Fellow, IEEE \qquad 
Guangtao Zhai$^{1,2}$, Senior Member, IEEE
\end{tabular}}


\address{$^{1}$Shanghai Jiao Tong University\quad$^{2}$Peng Cheng Laboratory\quad$^{3}$Toronto Metropolitan University}

\maketitle

\thispagestyle{fancy}
\fancyhead{}
\renewcommand{\headrulewidth}{0pt} 
\cfoot{\small{© 2024 IEEE. Personal use of this material is permitted. Permission from IEEE must be obtained for all other uses, in any current or future media, including reprinting/republishing this material for advertising or promotional purposes, creating new collective works, for resale or redistribution to servers or lists, or reuse of any copyrighted component of this work in other works.}}
\rfoot{}

\begin{abstract}
%

In the realm of media technology, digital humans have gained prominence due to rapid advancements in computer technology. However, the manual modeling and control required for the majority of digital humans pose significant obstacles to efficient development. The speech-driven methods offer a novel avenue for manipulating the mouth shape and expressions of digital humans. Despite the proliferation of driving methods, the quality of many generated talking head (TH) videos remains a concern, impacting user visual experiences. To tackle this issue, this paper introduces the Talking Head Quality Assessment (THQA) database, featuring 800 TH videos generated through 8 diverse speech-driven methods. Extensive experiments affirm the THQA database's richness in character and speech features. Subsequent subjective quality assessment experiments analyze correlations between scoring results and speech-driven methods, ages, and genders. In addition, experimental results show that mainstream image and video quality assessment methods have limitations for the THQA database, underscoring the imperative for further research to enhance TH video quality assessment. The THQA database is publicly accessible at https://github.com/zyj-2000/THQA.

\end{abstract}
\vspace{0.2cm}
\section{Introduction}
\label{sec:intro}

\begin{figure}[!t]
    \vspace{-0cm}
    \centering
    \includegraphics[width =0.95\linewidth]{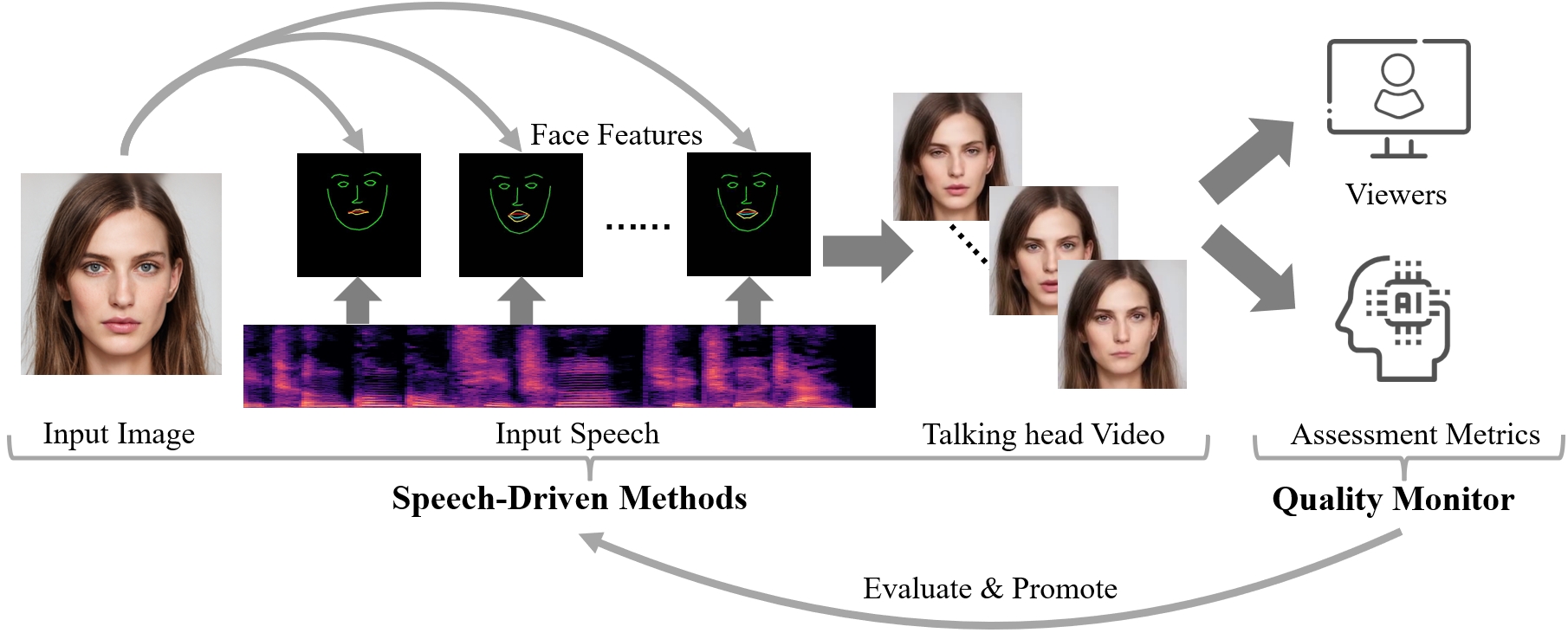}
    \vspace{-0.2cm}
    \caption{Motivation of our works. Quality assessment against talking head videos is essential for both the further development of speech-driven methods and the enhancement of the user experience.}
    \label{fig:motivation}
    \vspace{-0.2cm}
\end{figure}

\noindent The digital human, as an emerging form of digital media technology, has found extensive applications in diverse industries such as entertainment, medicine, film and television, owing to its capability to mimic or closely resemble human appearance, voice, and movements. However, the current landscape of digital human design and production remains arduous and time-consuming, predominantly reliant on manual efforts by skilled professionals. This manual design approach significantly hampers the efficiency of digital human content production, particularly concerning the intricate domain of head design and drive mechanisms. In response to this challenge, the advent and proliferation of artificial intelligence (AI) present a promising solution as shown in Fig.~\ref{fig:motivation}. Although speech-driven approaches have been successively proposed, simplifying the design of expressions and actions for digital human faces, there is still a lack of quality assessment metrics specifically for AI-generated videos of talking heads. These quality metrics can not only effectively assess the quality of talking head (TH) videos, but also indirectly contribute to the further development of speech-driven methods, which in turn can provide users with a higher-quality visual experience.

\begin{table*}[!htp]
\centering
\renewcommand\arraystretch{0.5}
\caption{The comparison of digital human quality assessment databases and proposed database.}
\resizebox{\linewidth}{!}{\begin{tabular}{ccccc}
\toprule
Database        & Subjects   & Subject-rated Models  &Content   & Description            \\
\midrule
DHH-QA \cite{dhhqa}    &55 & 1,540 & Textured Mesh  & Scanned Real Human Heads          \\
DDHQA \cite{ddhqa}   &2 &800  &  Textured Mesh   & Dynamic 3D Digital Human           \\
SJTU-H3D \cite{h3d} &40 &1,120    & Textured Mesh   & Static 3D Digital Human  \\
6G-DTQA \cite{6gqa}  &5 & 400  & Textured Mesh   &   Digital Twins Transmitted Under 6G Networks  \\
THQA(Proposed)           & 20  & 800 & AI-Generated Video  & Talking Heads Driven by AI\\
\bottomrule
\end{tabular}}
\vspace{-0.4cm}
\label{tab:databases}
\end{table*}

Unfortunately, prevailing methodologies in the evaluation of generated talking head videos  still adhere to a paradigm wherein a comparison with the original video is retained. Notably, the Frechet Inception Distance (FID) \cite{fid} and cosine similarity (CSIM) persist as the primary metrics employed for such assessments. However, the limitations of these metrics manifest in two fundamental aspects. Firstly, these objective assessment metrics focus solely on the image or video similarity, disregarding the holistic visual experience imparted to the viewer-user by the entire generated content. Secondly, their reliance on the original reference video poses a substantial constraint, given its inaccessibility to end-users, thereby severely limiting their applicability. While metrics like CPBD \cite{cpbd} and CGIQA \cite{cgiqa} have been incorporated in some recent works \cite{cumt} to gauge blur levels and aesthetic features, none of the widely utilized evaluation metrics are specifically tailored for TH videos. Addressing this challenge necessitates the development of a publicly accessible, large-scale database of TH videos at first. Consequently, this paper directs its focus towards the quality assessment of speech-driven solutions for THs, building the talking head quality assessment (THQA) database. The construction of THQA database involves the generation of 20 face images using StyleGAN \cite{stylegan}, and the subsequent head-driven manipulation of each image through 8 distinct speech-driven schemes, resulting in the creation of 800 TH videos. Subsequently, a subjective experiment is conducted on the THQA database to aggregate mean opinion scores (MOS) for each generated TH video. Finally, the performance of commonly employed quality assessment methods is verified, with the outcomes presented as a benchmark.

\section{Related Works}

\subsection{Speech-Driven Methods}
Speech-driven technology involves the application of AI to synchronize a character's facial features and mouth movements with the given speech, thereby creating the illusion of a TH. This technique can be categorized into 2D head speech driving and 3D head speech driving, with the former being more mature. Within the domain of 2D head driving, further classification can be made between person-dependent driving method \cite{suwajanakorn2017synthesizing} and person-independent driving method \cite{sadtalker,makelttalk,audio2head,iplap,dreamtalk,wav2lip,videoretalking,dinet}. The former necessitates tailored training for each driving entity, resulting in increased complexity and higher time overhead. Moreover, its applicability is challenged in scenarios lacking true speaking videos. Contrastingly, the person-independent approach addresses these challenges adeptly. It offers ease of use, high generalizability, and applicability to any driven subject, albeit with marginally reduced facial and movement details. This paper opts for the person-independent method in constructing the THQA database, given that the selected subjects are generated by StyleGAN, lacking corresponding speaking videos for individual subjects.

\subsection{Digital Human Quality Assessment}
With the escalating prevalence of digital humans in diverse facets of life and production, users are increasingly articulating elevated expectations regarding the quality and experiential aspects of digital humans. Consequently, the quality assessment of digital humans has garnered heightened attention. However, challenges persist within the domain of digital humans, such as prolonged design cycles and difficulties in amassing relevant digital human materials. These challenges present impediments to the advancement of quality assessment for digital humans. Encouragingly, as delineated in Table~\ref{tab:databases}, several pertinent databases have been established and applied to formulate objective quality assessment methods \cite{vitqa,zhang2023geometry,chen2023no,zhang2024reduced} for digital humans. In addition, some quality assessment methods \cite{kou2023stablevqa,dong2023light,wu2024towards} for 2D media also provide valuable ideas and inspiration for the field. A perusal of Table~\ref{tab:databases} reveals that while these established databases exhibit richness in content, there is a notable dearth of specificity concerning the TH. Therefore, the proposed THQA database in this paper aims to address this gap by further augmenting the existing database landscape for digital human quality assessment, thereby fostering advancements in the field of digital human quality evaluation.

\begin{table}[!tp]
    
    \centering
    \caption{Details of the selected human faces.}
    \begin{tabular}{c:c:c|c:c:c}
    \toprule
         ID & Gender & Age & ID & Gender & Age\\ \hline
         \#1 &Male &Child &\#11 &Female & Child \\
         \#2 &Male &Young &\#12 &Female &Child\\
         \#3 &Male &Young &\#13 &Female &Young\\
         \#4 &Male &Young &\#14 &Female &Young\\
         \#5 &Male &Middle &\#15 &Female &Young\\
         \#6 &Male &Middle &\#16 &Female &Middle\\
         \#7 &Male &Middle &\#17 &Female &Middle\\
         \#8 &Male &Old &\#18 &Female &Old\\
         \#9 &Male &Old &\#19 &Female &Old\\
         \#10 &Male &Old &\#20 &Female &Old\\
        
    \bottomrule
    \end{tabular}
    \label{tab:number}
    \vspace{-0.3cm}
\end{table}

\begin{figure}[!t]
    \vspace{-0cm}
    \centering
    \includegraphics[width =0.9\linewidth]{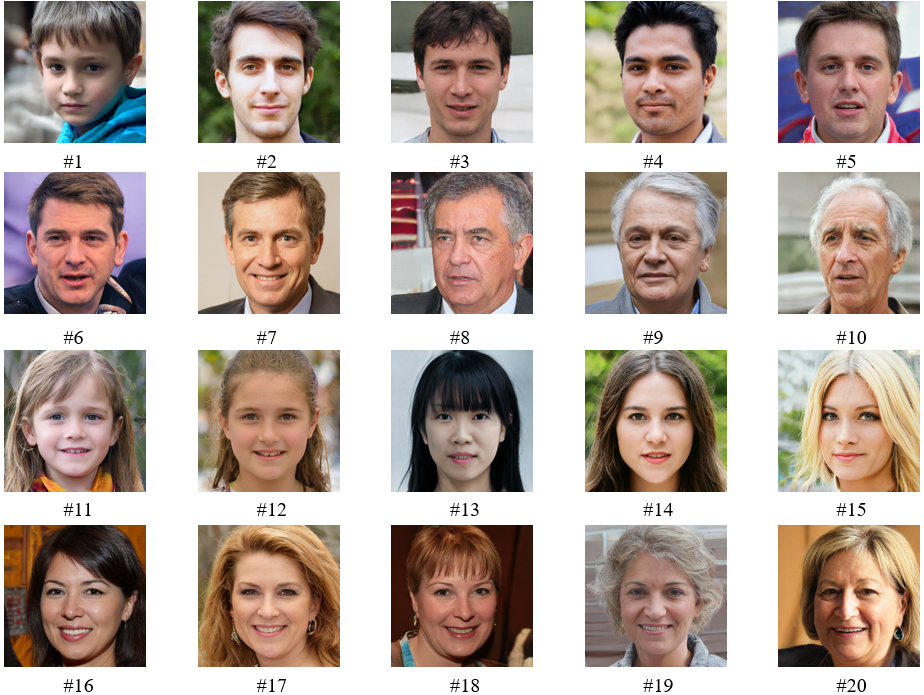}
    \vspace{-0.4cm}
    \caption{Overview of the selected human faces.}
    \label{fig:overview}
    \vspace{-0cm}
\end{figure}
\begin{figure}[!t]
    \centering
    \includegraphics[width = 1\linewidth]{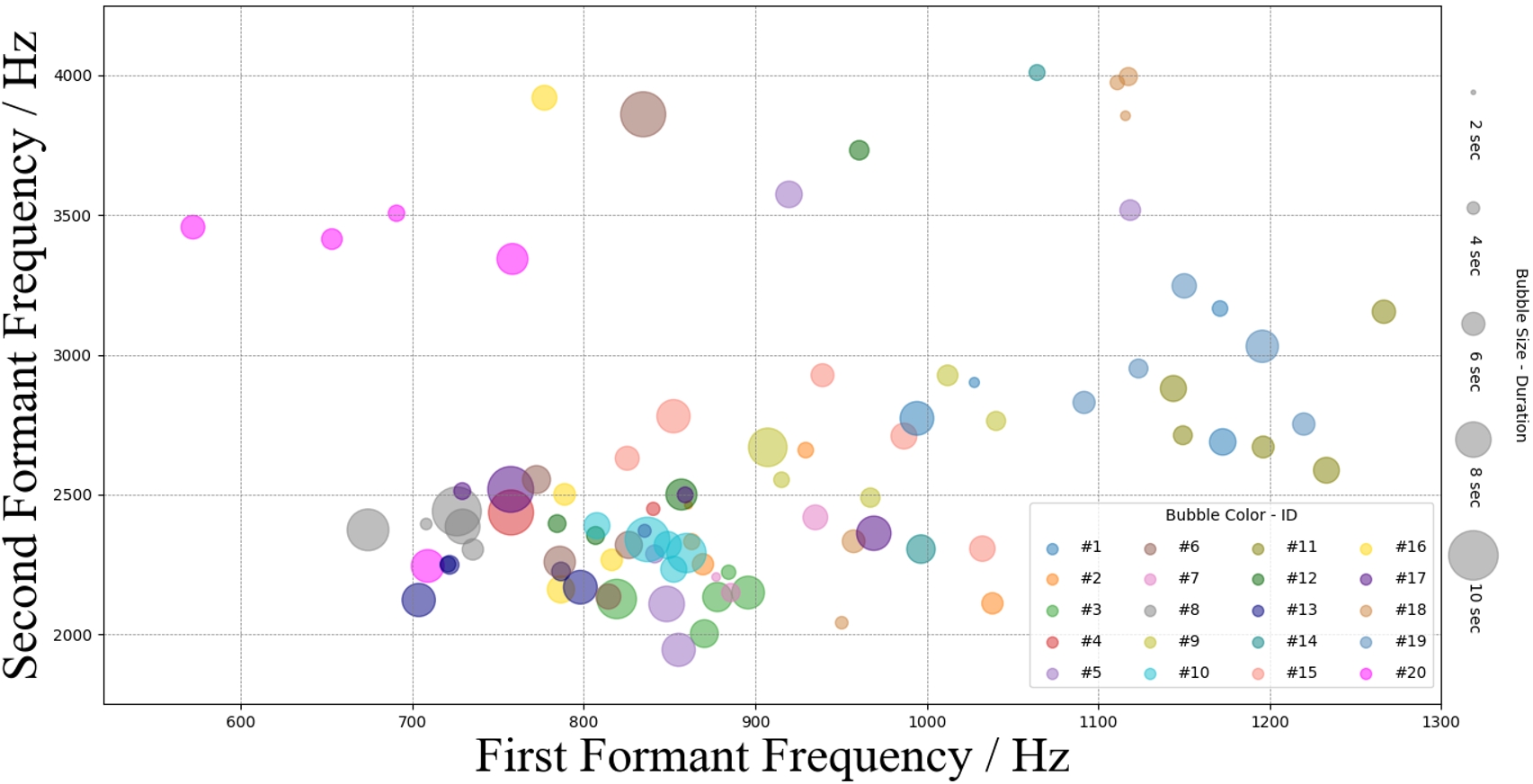}
    
    \caption{Phonological attributes of the selected speech, excluding samples displaying resonance peak merging. The first formant peak correlates with the amplitude of mouth opening and closing. Concurrently, the second formant peak is associated with tongue position. Bubble color denotes the subject ID corresponding to the speech, while bubble size serves as an indicator of speech duration.}
    \label{fig:audio}
    \vspace{-0.5cm}
\end{figure}

\begin{figure}[t]
    \vspace{-0.5cm}
    \centering
    \includegraphics[width = 1\linewidth]{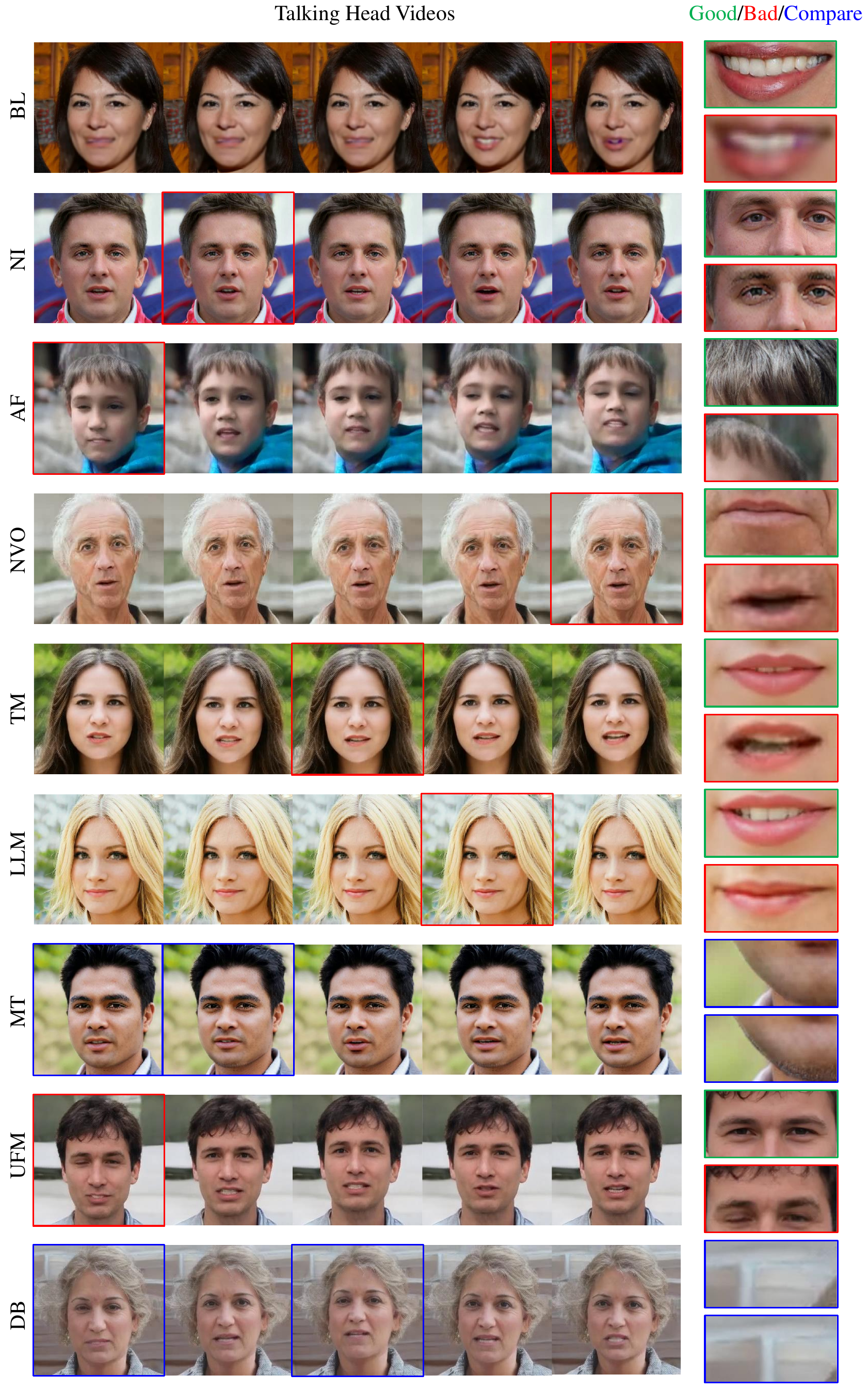}
    \vspace{-0.8cm}
    \caption{Diverse distortions observed in TH videos. The left side displays frames from the video containing distortions, while the right side presents a magnified image for localized examination. The green, red, and blue borders are employed to respectively delineate the categories of good samples, distortion samples, and frames for comparison between video frames.}
    \label{fig:distortions}
    \vspace{-0.2cm}
\end{figure}

\begin{figure}[!t]
    \vspace{-0.5cm}
    \centering
    \includegraphics[width=0.9\linewidth]{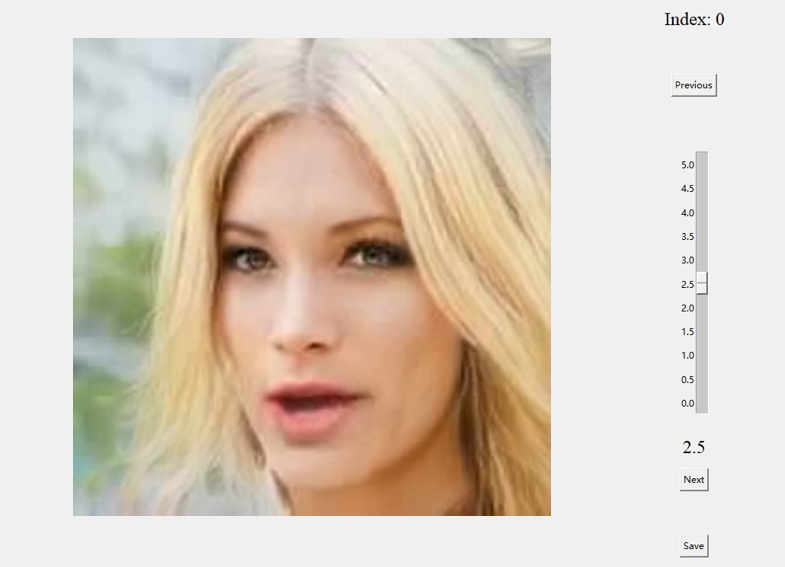}
    \vspace{0cm}
    \caption{The screenshot of the subjective quality assessment interface.}
    \label{fig:screen}
\end{figure}

\begin{figure}[!t]
    \centering
    \vspace{-0.2cm}
    
    \subfigure[]{\includegraphics[width=0.45\linewidth]{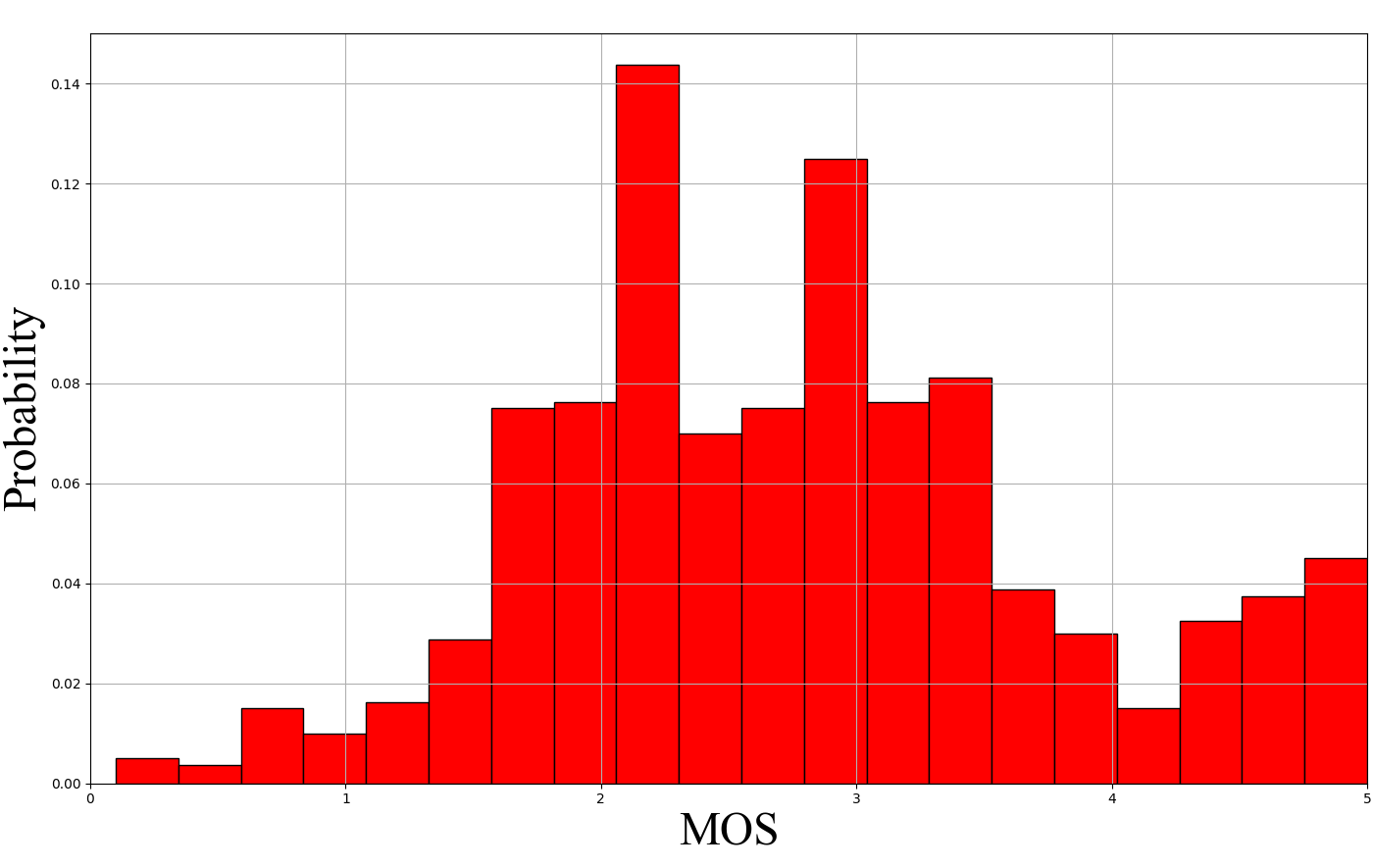}}
    \subfigure[]{\includegraphics[width=0.45\linewidth]{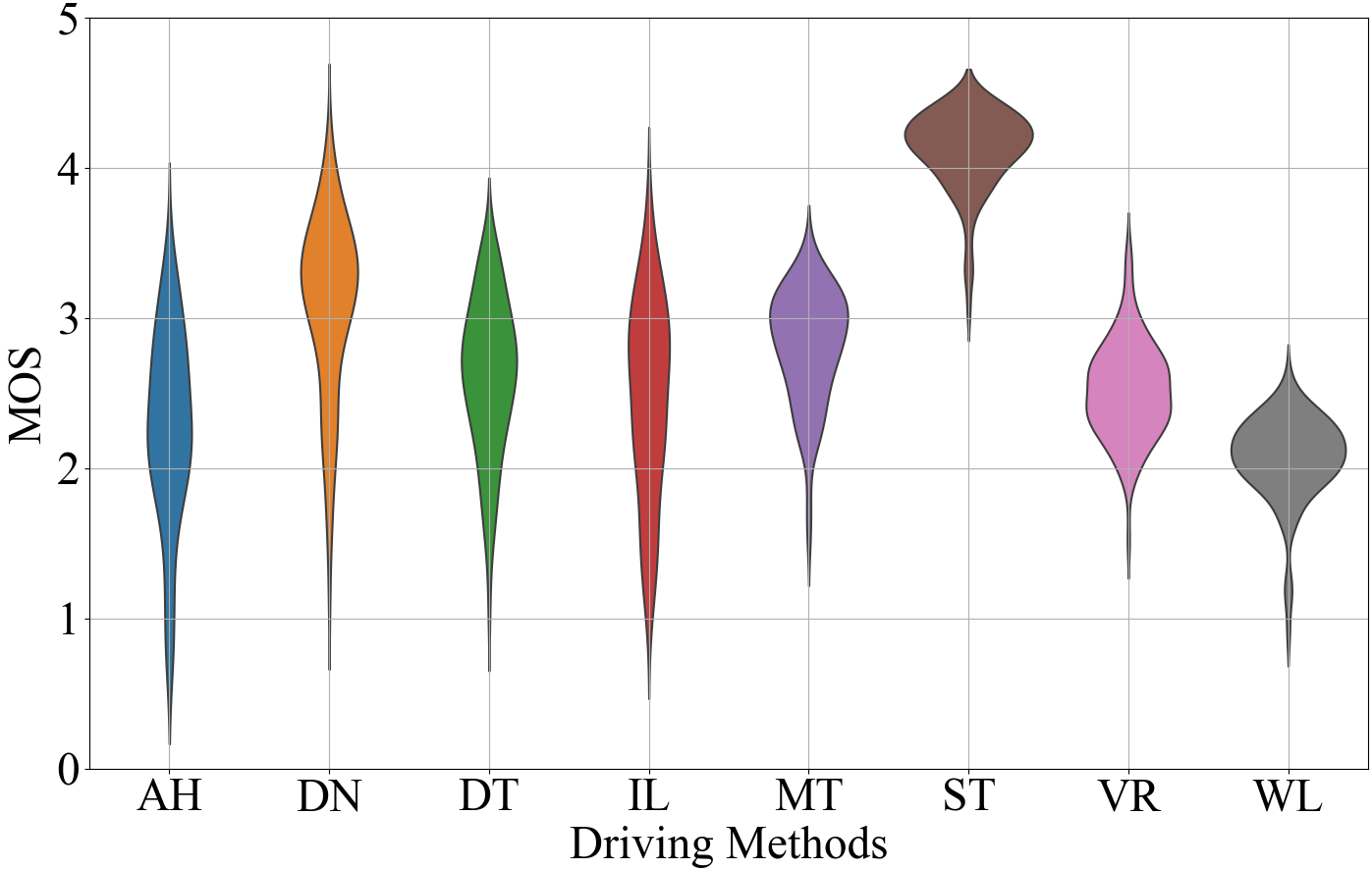}}
    
    \subfigure[]{\includegraphics[width=0.45\linewidth]{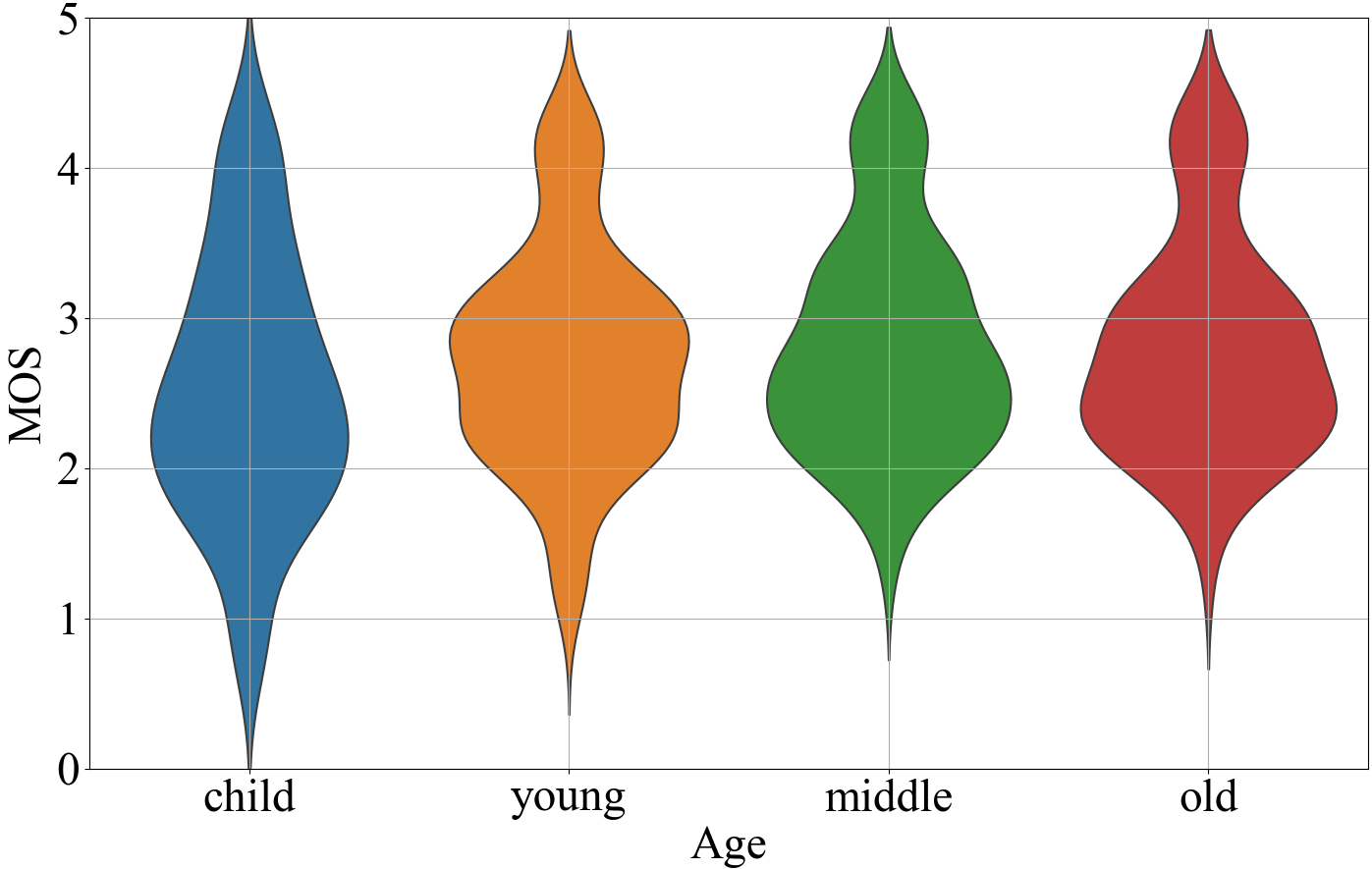}}
    \subfigure[]{\includegraphics[width=0.45\linewidth]{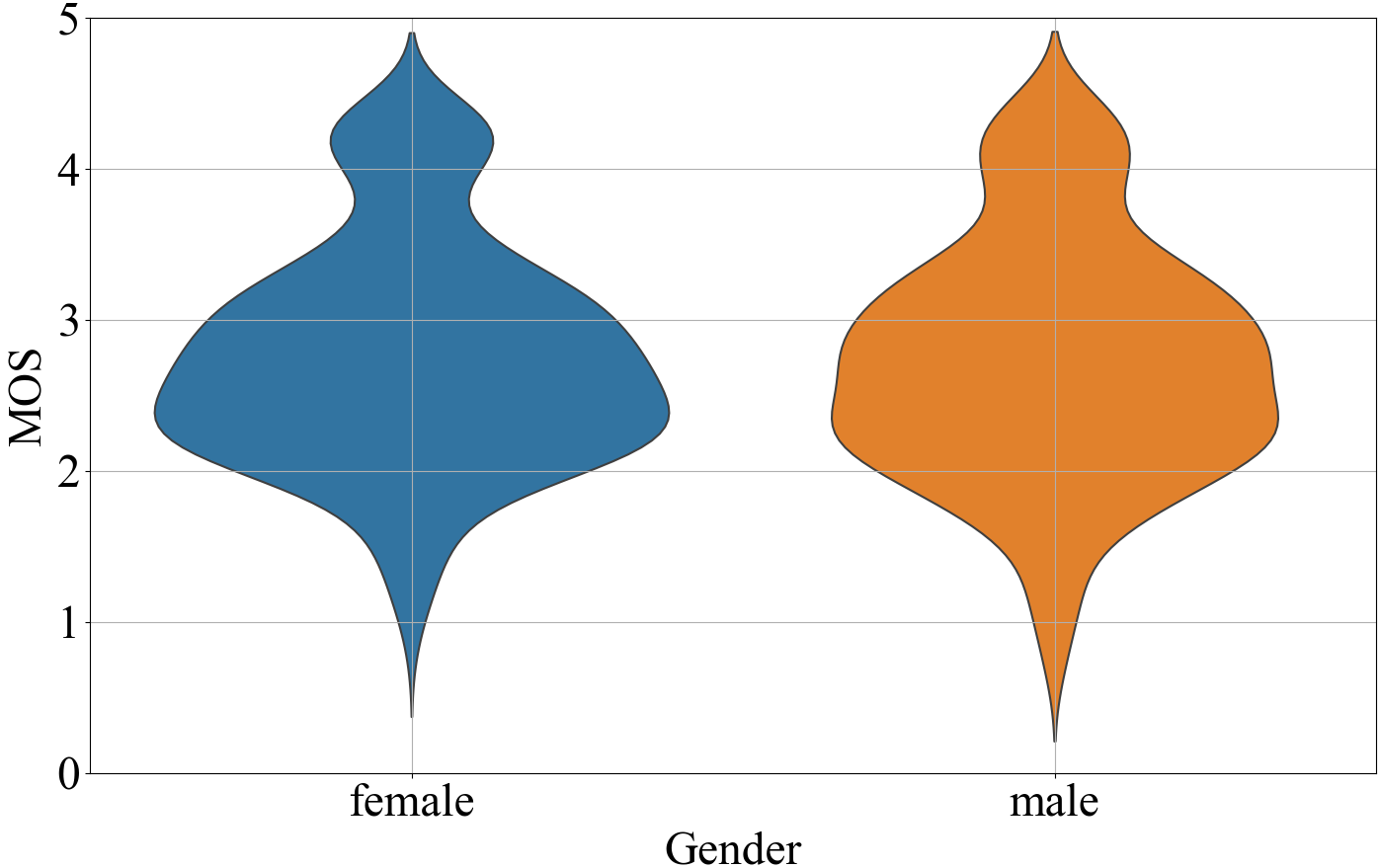}}
    
    \vspace{-0.2cm}
    \caption{Distributions of the MOSs.}
    \label{fig:distribution}
    \vspace{-0.2cm}
\end{figure}

\begin{table*}[!tp]
    \centering
    \caption{Details of the speech-driving methods employed for the generation of TH videos. }
    \resizebox{0.7\linewidth}{!}{\begin{tabular}{c|c|c|c|c|c}
    \toprule
         Type & Label &Methods & Year & Head Motion & Output Resolution\\ \hline
          \multirow{4}{*}{Image-based} &MT &MakeIttalk \cite{makelttalk} &2020 &\checkmark &256×256 \\  
          &AH&Auido2Head \cite{audio2head}&2021 &\checkmark &256×256 \\
          &ST&Sadtalker \cite{sadtalker}&2023 &\checkmark &512×512 \\
          &DT&Dreamtalk \cite{dreamtalk}&2023 &\checkmark &256×256  \\
           \hdashline
         \multirow{4}{*}{Video-based} &WL&Wav2Lip \cite{wav2lip}&2020 &\ding{53} &1024×1024 \\
         &VR&Video-Retalking \cite{videoretalking}&2022 &\ding{53} &1024×1024 \\
         &DN&DINet \cite{dinet}&2023 &\ding{53} &1024×1024 \\
         &IL&IP-LAP \cite{iplap}&2023 &\ding{53} &1024×1024 \\

    \bottomrule
    \end{tabular}}
    \label{tab:driving}
    \vspace{-0.3cm}
\end{table*}

\begin{table*}[!tp]
    
    \centering
    \caption{Distortions potentially manifesting in TH videos. Certain distortions are distinctive to a particular driving method or methods and exhibit universality. Additionally, specific distortions pertain exclusively to certain character images or particular speech segments, manifesting idiosyncratically.}
    \begin{tabular}{c|c|c|c}
    \toprule
         Type & Label & Distortion & Description\\ \hline
         \multirow{3}{*}{Image Quality} &BL &Blur & Local or global blurring in the video\\
         &NI &Noise &Noticeable noise in the video\\
         &AF &Artifact &Artifacts appearing at the edges of the face\\ \hdashline
         \multirow{3}{*}{Lip-sound Consistency} & NVO &No Voice but mouth Open &Characters still have lip shapes when they are voiceless\\
         & TM &Twisted Mouth &Characters show significant lip asymmetry\\ 
         & LLM &Little Lip Motion &The character's lips barely change when speaking\\ \hdashline
         \multirow{3}{*}{Overall Naturalness} & MT &Muscle Twitch &Noticeable facial muscle twitch in the video\\
         & UFM &Unnatural Facial Movements &Various unnatural expressions and gestures in the video\\
         & DB &Distorted Background &Noticeable change of background in the video\\
        
    \bottomrule
    \end{tabular}
    \label{tab:dis}
    \vspace{-0.3cm}
\end{table*}

\section{Database Construction}

\subsection{Source Materials Collection}
\noindent For the construction of the THQA database, a meticulous selection process is undertaken. Twenty images are chosen from the StyleGAN database\footnote{https://github.com/a312863063/generators-with-stylegan2}, as illustrated in Table~\ref{tab:number} and Fig.~\ref{fig:overview}. Notably, the selected generated face images encompass a balanced distribution, with 10 male and 10 female images. Additionally, the images represent various age groups, including child, young,  middle, and old age, ensuring universality. All selected images adhere to a resolution of 1024 × 1024.

Subsequently, for each selected character image, five speech segments are manually selected from Common Voicee\footnote{https://commonvoice.mozilla.org}. To enhance the realism and naturalness of the driving effect, the voices are chosen based on the gender and age of the selected image, with a concerted effort to maintain consistent timbre for the same image. In total, 100 = 5 × 20 are thoughtfully assigned to each face image. To ensure the selected speech segments encompass a broad range of phonemes and varied durations for comprehensive comparison across different speech-driving methods, the inverse spectral method is employed to estimate formant peaks and record the duration of each speech segment. The outcomes of this estimation are depicted in Fig.~\ref{fig:audio}. Analysis of Fig. ~\ref{fig:audio} reveals several significant findings: 1) The selected speeches exhibit a diverse range of first and second formant peak frequencies (650-1300 Hz and 2000-4000 Hz), indicating a rich variety of mouth shapes and tongue positions. This substantiates the generalizability of the proposed database. 2) The phonological features of the 5 speeches assigned to the same subject demonstrate similarity, underscoring a cohesive match between the characters and their assigned speeches. 3) All recorded speeches have durations ranging from 3-9 seconds, contributing to the enrichment of phonological features within the proposed database.

\subsection{Human Head Driving \& Quality Analysis}
The amalgamation of character face images and speech underpins the implementation of a speech-driven methodology for the generation of TH videos. This study employs 8 prominent speech-driving methods, as delineated in Table~\ref{tab:driving}. Notably, four image-based speech-driven approaches and four video-based speech-driven approaches are specifically chosen for the generation of TH videos. It is noteworthy that the original code, as provided by the respective authors, is utilized for all methods, resulting in varying resolutions for the generated TH videos. It is pertinent to mention that the video-based methods alter the mouth shape in accordance with the input speech while preserving the original head movements. Conversely, as the selected character images are static, video synthesis is achieved by replicating these images. Although this synthesized video introduces a temporal dimension to the images, it merely retains the resolution of the original images and does not encompass the generation of head movements.

Subsequent to video generation, the TH videos are scrutinized from three dimensions: image quality, lip-sound consistency, and overall naturalness. The evaluation reveals the occurrence of nine common quality distortions, enumerated in Table~\ref{tab:dis}. To provide a more visually intuitive representation of these distortions, a typical TH video is selected for each distortion, as depicted in Fig.~\ref{fig:distortions}. The conjunction of the distortions illustrated in Fig. ~\ref{fig:distortions}, underscores the prevailing quality issues inherent in speech-driven generated TH videos. This reiterates the imperative need for a comprehensive quality assessment framework for TH videos.

\subsection{Subjective Experiment}
In accordance with the ITU-R BT.500-13 guidelines \cite{bt2002methodology}, subjective experiments are conducted in a controlled laboratory environment. The talking head videos generated for evaluation were randomly presented on iMac monitors with a resolution of 4096 $\times$ 2304. The subjective quality assessment interface utilized in the experiments is depicted in Fig.~\ref{fig:screen}.

A cohort of 40 participants, consisting of 20 males and 20 females, are systematically invited to participate in the subjective experiment. The participants are positioned at a distance from the screen equivalent to twice the screen height within an indoor setting with standardized illumination levels. Before the initiation of the subjective experiment, an instructional session is conducted to familiarize the subjects with the nuances of the quality assessment task. The entire subjective experiment is partitioned into 16 sessions, each featuring the presentation of 50 generated TH videos. Interspersed between these sessions are 30-minute breaks, and participants are constrained to engage in no more than 4 sessions within a single day. Within each session, the TH video is played once, enabling participants to rate the quality on a scale from 0 to 5, with a minimum rating interval of 0.1. Stringent measures are enforced to ensure that each TH video underwent evaluation by all 40 participants, culminating in the accumulation of a comprehensive database comprising 32,000 subjective ratings.

\subsection{Subjective Data Processing}
Based on prior research \cite{li2023agiqa,cgiqa} , z-scores are employed to perform data processing on the collected subjective ratings and the process can be represented as follow:

\begin{equation}
z_{i j}=\frac{r_{i j}-\mu_{i}}{\sigma_{i}},
\end{equation}
where $r_{ij}$ denotes the quality rating provided by the $i$-th subject on the $j$-th TH video, $\mu_{i}=\frac{1}{N_{i}} \sum_{j=1}^{N_{i}} r_{i j}$, $\sigma_{i}=\sqrt{\frac{1}{N_{i}-1} \sum_{j=1}^{N_{i}}\left(r_{i j}-\mu_{i}\right)}$, and $N_i$ is the number of TH videos assessed by subject $i$. Furthermore, quality ratings from unreliable subjects are discarded following the subject rejection procedure recommended in \cite{bt2002methodology}.
Subsequently, the obtained z-scores undergo linear rescaling to the range $[0,5]$. The MOS of TH video $j$ is then computed by averaging the rescaled z-scores: 

\begin{equation}
M O S_{j}=\frac{1}{M} \sum_{i=1}^{M} z_{i j}^{'},
\end{equation}
where $M O S_{j}$ signifies the MOS for the $j$-th TH video, $M$ denotes the number of the valid subjects, and $z_{i j}^{'}$ represents the rescaled z-scores.

\subsection{Subjective Data Analysis}
We further plot the distributions of the MOSs, which are shown in Fig. \ref{fig:distribution}(a). Additionally, we conduct a comprehensive analysis of subjective ratings based on three distinct perspectives: driving methods, age, and gender, as illustrated in Fig.~\ref{fig:distribution}(b)(c)(d), respectively. Scrutiny of Fig.~\ref{fig:distribution} facilitates several observations. 1) The overall MOS distribution exhibits a bell-shaped distribution, with fewer instances at both extremes and a higher probability in the middle. This trend indicates a prevalence of TH videos with low to medium quality in the majority, reinforcing the inclusivity of various quality levels within the proposed THQA database, rendering it suitable for comprehensive THQA. 2) Among the eight selected speech-driven methods, Sadtalker stands out by consistently maintaining a high level of stability in the quality of generated talking head videos. In contrast, the quality of videos generated by other methods appears to be more susceptible to the influence of driving voice and character image. 3) Analysis across different age groups reveals challenges in driving methods for children. This difficulty may stem from the training datasets predominantly comprising adult data, leading to a scarcity of child-specific data. Consequently, the generalization of driving approaches to the children age group is compromised. 4) Subjective scoring results across different genders exhibit similarity, suggesting that gender is not a significant factor influencing the quality assessment of TH videos.

\section{Benchmark Experiment}
\subsection{Competitors}
In light of the absence of a reference video for the generated TH video, we employ no-reference (NR) image quality assessment (IQA) and video quality assessment (VQA) methods for conducting performance testing experiments. Specifically, NR IQA methods encompass BRISQUE \cite{brisque}, NIQE \cite{niqe}, IL-NIQE \cite{ilniqe}, and CPBD \cite{cpbd}. Meanwhile, NR VQA methods include VIIDEO \cite{viideo}, V-BLIINDS \cite{vblinds}, TLVQM \cite{tlvqm}, VIDEVAL \cite{videval}, VSFA \cite{vsfa}, RAPIQUE \cite{rapique}, SimpVQA \cite{simpvqa}, FAST-VQA \cite{fastvqa}, and BVQA \cite{bvqa}, with VSFA, RAPIQUE, SimpVQA, FAST-VQA, and BVQA being deep learning-based methods. Notably, all methods utilize the source code provided by the respective authors and adhere to the original parameter settings. It is essential to underscore that for the IQA algorithm, the TH video is sampled at a rate of one frame per second for quality assessment. The final video quality is derived by averaging the quality scores obtained from all the extracted frames.
\begin{table}[t]\small
\renewcommand\tabcolsep{4.2pt}
  \caption{Benchmark performance on the THQA database. Best in \bf{bold}. }
  \vspace{-0.2cm}
  \begin{center}
  \begin{tabular}{c|c|cccc}
    \toprule
    Type &  Method & SRCC$\uparrow$ &  PLCC$\uparrow$ & KRCC$\uparrow$ & RMSE$\downarrow$ \\ \hline
    \multirow{4}{*}{IQA} 
    &BRISQUE  & 0.4856 & 0.5970& 0.3454&0.8227\\
    &NIQE  & 0.1007& 0.1389& 0.0804&0.9673\\
    &IL-NIQE  & 0.1199& 0.1298& 0.0979&0.9691\\
    &CPBD  & 0.1184& 0.1802& 0.0932&0.9532\\\hdashline
    \multirow{9}{*}{VQA} 
    &VIIDEO & 0.1777 &0.1891 & 0.1354& 0.9595\\
    &V-BLIINDS & 0.4949 & 0.6403&0.3533&0.7976\\
    &TLVQM & 0.0254 &0.0355& 0.0209&1.0853\\
    &VIDEVAL & 0.0317 & 0.0358& 0.0231&1.1916\\
    &VSFA & \bf{0.7601} & \bf{0.8106}& \bf{0.5830}&\bf{0.5966}\\
    &RAPIQUE & 0.1789 & 0.1908& 0.1277&1.0162\\
    &SimpVQA & 0.6800 & 0.7592 & 0.5052 & 0.6361\\
    &FAST-VQA & 0.6389 & 0.7441& 0.4677&0.6983\\
    &BVQA & 0.7287 & 0.7985&0.5549 &0.6094\\
    \bottomrule
  \end{tabular}
  \end{center}
  \vspace{-1cm}
  \label{tab:performance}
\end{table}
\subsection{Experimental Setup}
The THQA database is systematically partitioned into 5 distinct groups. Each group consists of 160 TH videos generated from 4 distinct images, ensuring a balanced distribution and meticulous measures are implemented to preclude any content overlap between groups. For the testing of various quality assessment algorithms, a 5-fold cross-validation approach is adopted and the average performance metrics are reported. To quantitatively analyze the efficacy of each method, four widely recognized quality assessment metrics are employed: Spearman Rank Correlation Coefficient (SRCC), Kendall’s Rank Correlation Coefficient (KRCC), Pearson Linear Correlation Coefficient (PLCC), and Root Mean Squared Error (RMSE). The evaluation metrics SRCC, PLCC, and KRCC yield values within the range of 0 to 1, with higher proximity to 1 indicating a better alignment with human visual perception. In the case of RMSE, a closer proximity to 0 signifies increased effectiveness of the method.

\subsection{Performance Analysis}
The experimental results are presented in Table~\ref{tab:performance}, offering insights that lead to several noteworthy conclusions. 1) Notably, deep learning-based quality assessment methods, except RAPIQUE, exhibit superior performance compared to traditional methods reliant on manually extracted features. This superiority can be attributed to the inherent challenge of employing specific features for the quality assessment of AI-generated videos such as those in the TH domain. 2) VSFA emerges as the leading performer among all the compared methods for TH videos. However, it is important to note that despite achieving state-of-the-art performance, the evaluation results for TH videos still exhibit a discernible gap when compared to subjective human visual perception. 3) The performance evaluation of existing quality assessment methods on the THQA database underscores the limitations inherent in these methods for TH videos. Consequently, it is evident that more effective and accurate assessment algorithms need to be explored and developed to address these limitations.

\section{Conclusion}
In this paper, we have constructed a quality assessment database for talking heads (TH) videos named THQA. This database encompasses 800 TH videos generated through 8 distinct speech-driven methods. Our analysis involves a thorough examination of the collected images, speech data, and the resultant generated videos. Furthermore, we conduct subjective scoring experiments to validate the representativeness of THQA, affirming its utility as a guiding framework for the quality assessment of TH videos. In conclusion, we perform a comparative evaluation of the performance of various mainstream assessment methods based on THQA database. The outcomes reveal that the majority of existing assessment methods exhibit limitations in effectively assessing the quality of TH videos.
\bibliographystyle{IEEEbib}
\bibliography{icme2022template}

\begin{thebibliography}{10}

\bibitem{dhhqa}
Zicheng Zhang, Yingjie Zhou, Wei Sun, Xiongkuo Min, Yuzhe Wu, and Guangtao Zhai,
\newblock ``Perceptual quality assessment for digital human heads,''
\newblock in {\em ICASSP}. IEEE, 2023, pp. 1--5.

\bibitem{ddhqa}
Zicheng Zhang, Yingjie Zhou, Wei Sun, Wei Lu, Xiongkuo Min, Yu~Wang, and Guangtao Zhai,
\newblock ``Ddh-qa: A dynamic digital humans quality assessment database,''
\newblock in {\em ICME}. IEEE, 2023, pp. 2519--2524.

\bibitem{h3d}
Zicheng Zhang, Wei Sun, Yingjie Zhou, Haoning Wu, Chunyi Li, Xiongkuo Min, Xiaohong Liu, Guangtao Zhai, and Weisi Lin,
\newblock ``Advancing zero-shot digital human quality assessment through text-prompted evaluation,''
\newblock {\em arXiv preprint arXiv:2307.02808}, 2023.

\bibitem{6gqa}
Zicheng Zhang, Yingjie Zhou, Long Teng, Wei Sun, Chunyi Li, Xiongkuo Min, Xiao-Ping Zhang, and Guangtao Zhai,
\newblock ``Quality-of-experience evaluation for digital twins in 6g network environments,''
\newblock {\em IEEE Transactions on Broadcasting}, 2024.

\bibitem{fid}
Martin Heusel, Hubert Ramsauer, Thomas Unterthiner, Bernhard Nessler, and Sepp Hochreiter,
\newblock ``Gans trained by a two time-scale update rule converge to a local nash equilibrium,''
\newblock {\em Advances in neural information processing systems}, vol. 30, 2017.

\bibitem{cpbd}
Niranjan~D Narvekar and Lina~J Karam,
\newblock ``A no-reference image blur metric based on the cumulative probability of blur detection (cpbd),''
\newblock {\em IEEE TIP}, vol. 20, no. 9, pp. 2678--2683, 2011.

\bibitem{cgiqa}
Zicheng Zhang, Wei Sun, Yingjie Zhou, Jun Jia, Zhichao Zhang, Jing Liu, Xiongkuo Min, and Guangtao Zhai,
\newblock ``Subjective and objective quality assessment for in-the-wild computer graphics images,''
\newblock {\em ACM Transactions on Multimedia Computing, Communications and Applications}, vol. 20, no. 4, pp. 1--22, 2023.

\bibitem{cumt}
Yingjie Zhou, Yaodong Chen, Kaiyue Bi, Lian Xiong, and Hui Liu,
\newblock ``An implementation of multimodal fusion system for intelligent digital human generation,''
\newblock {\em arXiv preprint arXiv:2310.20251}, 2023.

\bibitem{stylegan}
Tero Karras, Samuli Laine, and Timo Aila,
\newblock ``A style-based generator architecture for generative adversarial networks,''
\newblock in {\em IEEE/CVF CVPR}, 2019, pp. 4401--4410.

\bibitem{suwajanakorn2017synthesizing}
Supasorn Suwajanakorn, Steven~M Seitz, and Ira Kemelmacher-Shlizerman,
\newblock ``Synthesizing obama: learning lip sync from audio,''
\newblock {\em ACM ToG}, vol. 36, no. 4, pp. 1--13, 2017.

\bibitem{sadtalker}
Wenxuan Zhang, Xiaodong Cun, Xuan Wang, Yong Zhang, Xi~Shen, Yu~Guo, Ying Shan, and Fei Wang,
\newblock ``Sadtalker: Learning realistic 3d motion coefficients for stylized audio-driven single image talking face animation,''
\newblock in {\em IEEE/CVF CVPR}, 2023, pp. 8652--8661.

\bibitem{makelttalk}
Yang Zhou, Xintong Han, Eli Shechtman, Jose Echevarria, Evangelos Kalogerakis, and Dingzeyu Li,
\newblock ``Makelttalk: speaker-aware talking-head animation,''
\newblock {\em ACM TOG}, vol. 39, no. 6, pp. 1--15, 2020.

\bibitem{audio2head}
Suzhen Wang, Lincheng Li, Yu~Ding, Changjie Fan, and Xin Yu,
\newblock ``Audio2head: Audio-driven one-shot talking-head generation with natural head motion,''
\newblock {\em arXiv preprint arXiv:2107.09293}, 2021.

\bibitem{iplap}
Weizhi Zhong, Chaowei Fang, Yinqi Cai, Pengxu Wei, Gangming Zhao, Liang Lin, and Guanbin Li,
\newblock ``Identity-preserving talking face generation with landmark and appearance priors,''
\newblock in {\em IEEE/CVF CVPR}, June 2023, pp. 9729--9738.

\bibitem{dreamtalk}
Yifeng Ma, Shiwei Zhang, Jiayu Wang, Xiang Wang, Yingya Zhang, and Zhidong Deng,
\newblock ``Dreamtalk: When expressive talking head generation meets diffusion probabilistic models,''
\newblock {\em arXiv preprint arXiv:2312.09767}, 2023.

\bibitem{wav2lip}
KR~Prajwal, Rudrabha Mukhopadhyay, Vinay~P Namboodiri, and CV~Jawahar,
\newblock ``A lip sync expert is all you need for speech to lip generation in the wild,''
\newblock in {\em Proceedings of the 28th ACM international conference on multimedia}, 2020, pp. 484--492.

\bibitem{videoretalking}
Kun Cheng, Xiaodong Cun, Yong Zhang, Menghan Xia, Fei Yin, Mingrui Zhu, Xuan Wang, Jue Wang, and Nannan Wang,
\newblock ``Videoretalking: Audio-based lip synchronization for talking head video editing in the wild,''
\newblock in {\em SIGGRAPH Asia 2022}, 2022, pp. 1--9.

\bibitem{dinet}
Zhimeng Zhang, Zhipeng Hu, Wenjin Deng, Changjie Fan, Tangjie Lv, and Yu~Ding,
\newblock ``Dinet: Deformation inpainting network for realistic face visually dubbing on high resolution video,''
\newblock {\em arXiv preprint arXiv:2303.03988}, 2023.

\bibitem{vitqa}
Yingjie Zhou, Zicheng Zhang, Wei Sun, Xiongkuo Min, Xianghe Ma, and Guangtao Zhai,
\newblock ``A no-reference quality assessment method for digital human head,''
\newblock in {\em ICIP}. IEEE, 2023, pp. 36--40.

\bibitem{zhang2023geometry}
Zicheng Zhang, Yingjie Zhou, Wei Sun, Xiongkuo Min, and Guangtao Zhai,
\newblock ``Geometry-aware video quality assessment for dynamic digital human,''
\newblock in {\em ICIP}. IEEE, 2023, pp. 1365--1369.

\bibitem{chen2023no}
Shi Chen, Zicheng Zhang, Yingjie Zhou, Wei Sun, and Xiongkuo Min,
\newblock ``A no-reference quality assessment metric for dynamic 3d digital human,''
\newblock {\em Displays}, vol. 80, pp. 102540, 2023.

\bibitem{zhang2024reduced}
Zicheng Zhang, Yingjie Zhou, Chunyi Li, Kang Fu, Wei Sun, Xiaohong Liu, Xiongkuo Min, and Guangtao Zhai,
\newblock ``A reduced-reference quality assessment metric for textured mesh digital humans,''
\newblock in {\em ICASSP 2024-2024 IEEE International Conference on Acoustics, Speech and Signal Processing (ICASSP)}. IEEE, 2024, pp. 2965--2969.

\bibitem{kou2023stablevqa}
Tengchuan Kou, Xiaohong Liu, Wei Sun, Jun Jia, Xiongkuo Min, Guangtao Zhai, and Ning Liu,
\newblock ``Stablevqa: A deep no-reference quality assessment model for video stability,''
\newblock in {\em Proceedings of the 31st ACM International Conference on Multimedia}, 2023, pp. 1066--1076.

\bibitem{dong2023light}
Yunlong Dong, Xiaohong Liu, Yixuan Gao, Xunchu Zhou, Tao Tan, and Guangtao Zhai,
\newblock ``Light-vqa: A multi-dimensional quality assessment model for low-light video enhancement,''
\newblock in {\em Proceedings of the 31st ACM International Conference on Multimedia}, 2023, pp. 1088--1097.

\bibitem{wu2024towards}
Haoning Wu, Hanwei Zhu, Zicheng Zhang, Erli Zhang, Chaofeng Chen, Liang Liao, Chunyi Li, Annan Wang, Wenxiu Sun, Qiong Yan, et~al.,
\newblock ``Towards open-ended visual quality comparison,''
\newblock {\em arXiv preprint arXiv:2402.16641}, 2024.

\bibitem{bt2002methodology}
RECOMMENDATION ITU-R BT,
\newblock ``Methodology for the subjective assessment of the quality of television pictures,''
\newblock {\em International Telecommunication Union}, 2002.

\bibitem{li2023agiqa}
Chunyi Li, Zicheng Zhang, Haoning Wu, Wei Sun, Xiongkuo Min, Xiaohong Liu, Guangtao Zhai, and Weisi Lin,
\newblock ``Agiqa-3k: An open database for ai-generated image quality assessment,''
\newblock {\em IEEE Transactions on Circuits and Systems for Video Technology}, 2023.

\bibitem{brisque}
Anish Mittal, Anush~Krishna Moorthy, and Alan~Conrad Bovik,
\newblock ``No-reference image quality assessment in the spatial domain,''
\newblock {\em IEEE TIP}, vol. 21, no. 12, pp. 4695--4708, 2012.

\bibitem{niqe}
Anish Mittal, Rajiv Soundararajan, and Alan~C Bovik,
\newblock ``Making a “completely blind” image quality analyzer,''
\newblock {\em IEEE SPL}, vol. 20, no. 3, pp. 209--212, 2012.

\bibitem{ilniqe}
Lin Zhang, Lei Zhang, and Alan~C Bovik,
\newblock ``A feature-enriched completely blind image quality evaluator,''
\newblock {\em IEEE TIP}, vol. 24, no. 8, pp. 2579--2591, 2015.

\bibitem{viideo}
Anish Mittal, Michele~A Saad, and Alan~C Bovik,
\newblock ``A completely blind video integrity oracle,''
\newblock {\em IEEE TIP}, vol. 25, no. 1, pp. 289--300, 2015.

\bibitem{vblinds}
Michele~A Saad, Alan~C Bovik, and Christophe Charrier,
\newblock ``Blind prediction of natural video quality,''
\newblock {\em IEEE TIP}, vol. 23, no. 3, pp. 1352--1365, 2014.

\bibitem{tlvqm}
Jari Korhonen,
\newblock ``Two-level approach for no-reference consumer video quality assessment,''
\newblock {\em IEEE TIP}, vol. 28, no. 12, pp. 5923--5938, 2019.

\bibitem{videval}
Zhengzhong Tu, Yilin Wang, Neil Birkbeck, Balu Adsumilli, and Alan~C Bovik,
\newblock ``Ugc-vqa: Benchmarking blind video quality assessment for user generated content,''
\newblock {\em IEEE TIP}, vol. 30, pp. 4449--4464, 2021.

\bibitem{vsfa}
Dingquan Li, Tingting Jiang, and Ming Jiang,
\newblock ``Quality assessment of in-the-wild videos,''
\newblock in {\em ACM MM}, 2019, pp. 2351--2359.

\bibitem{rapique}
Zhengzhong Tu, Xiangxu Yu, Yilin Wang, Neil Birkbeck, Balu Adsumilli, and Alan~C Bovik,
\newblock ``Rapique: Rapid and accurate video quality prediction of user generated content,''
\newblock {\em IEEE DOAJ}, vol. 2, pp. 425--440, 2021.

\bibitem{simpvqa}
Wei Sun, Xiongkuo Min, Wei Lu, and Guangtao Zhai,
\newblock ``A deep learning based no-reference quality assessment model for ugc videos,''
\newblock in {\em ACM MM}, 2022.

\bibitem{fastvqa}
Haoning Wu, Chaofeng Chen, Jingwen Hou, Liang Liao, Annan Wang, Wenxiu Sun, Qiong Yan, and Weisi Lin,
\newblock ``Fast-vqa: Efficient end-to-end video quality assessment with fragment sampling,''
\newblock in {\em ECCV}. 2022, pp. 538--554, Springer.

\bibitem{bvqa}
Bowen Li, Weixia Zhang, Meng Tian, Guangtao Zhai, and Xianpei Wang,
\newblock ``Blindly assess quality of in-the-wild videos via quality-aware pre-training and motion perception,''
\newblock {\em IEEE TCSVT}, vol. 32, no. 9, pp. 5944--5958, 2022.

\end{thebibliography}

\end{document}